\documentclass[runningheads]{llncs}
\usepackage{graphicx}
\usepackage{amsmath,amssymb} 
\usepackage{color}
\usepackage{algorithm}
\usepackage{hyperref}
\usepackage[noend]{algpseudocode}
\makeatletter
\def\BState{\State\hskip-\ALG@thistlm}
\makeatother

\begin{document}
\title{Cricket \textit{stroke} extraction: Towards creation of a large-scale cricket actions dataset} 

\titlerunning{Cricket \textit{stroke} extraction}
%
\author{Arpan Gupta\orcidID{0000-0002-9417-3169} \and
Sakthi Balan M.\orcidID{0000-0003-1817-7173}}
%
\authorrunning{A. Gupta and S. Balan}
%

\institute{The LNM Institute of Information Technology, Jaipur, Rajasthan, India. 
\email{\{arpan,sakthi.balan\}@lnmiit.ac.in}\\
\url{https://www.lnmiit.ac.in/}
}
\maketitle              
\begin{abstract}
In this paper, we deal with the problem of temporal action localization for a large-scale untrimmed cricket videos dataset. Our action of interest for cricket videos is a cricket \textit{stroke} played by a batsman, which is, usually, covered by cameras placed at the stands of the cricket ground at both ends of the cricket pitch. After applying a sequence of preprocessing steps, we have $\approx 73$ million frames for $1110$ videos in the dataset at constant frame rate and resolution. The method of localization is a generalized one which applies a trained random forest model for CUTs detection(using summed up grayscale histogram difference features) and two linear SVM camera models(CAM1 and CAM2) for first frame detection, trained on HOG features of CAM1 and CAM2 video shots. CAM1 and CAM2 are assumed to be part of the cricket \textit{stroke}. At the predicted boundary positions, the HOG features of the first frames are computed and a simple algorithm was used to combine the positively predicted camera shots. In order to make the process as generic as possible, we did not consider any domain specific knowledge, such as tracking or specific shape and motion features.

The detailed analysis of our methodology is provided along with the metrics used for evaluation of individual models, and the final predicted segments. We achieved a weighted mean TIoU of \textbf{$0.5097$} over a small sample of the test set.

\keywords{Cricket \textit{stroke} extraction \and sports video processing \and HOG \and shot boundary detection \and untrimmed videos \and temporal localization}
\end{abstract}

\section{Introduction}

The vision researchers are still a long way from achieving human level understanding of videos by a machine. Though, we get good results for basic tasks on images, but extending the same methods for videos may not be that straight-forward. Major challenges in understanding of videos include camera motions, illumination changes, partial occlusion etc. Many of these challenges can be avoided in applications where a stationary camera is used, since, the moving foreground objects are much easier to segment out. On the other hand, video content from a non-stationary camera, like telecast videos, movies and ego-centric videos, give rise to all these challenges. It is tough to  come up with a unified model that deals with all of the above challenges, while at the same time ensuring the real-time processing of the video frames. 

Vision researchers, after seeing the success of deep neural networks\cite{Krizhevsky2012} on ImageNet\cite{Russakovsky2015}, have tried to apply them for activity recognition tasks\cite{Tran2015,Karpathy2014,Simonyan2014,Baccouche2011}. The need for large-scale annotated video datasets, for training the deep neural networks, was a driving factor, that resulted in a number of such benchmark video datasets. Some of them are Sports-1M\cite{Karpathy2014}, Youtube-8M\cite{Abu-El-Haija2016}, ActivityNet\cite{Heilbron2015}, Kinetics\cite{Kay2017} etc.   

Activity recognition in sports telecast videos is an active area of research. There have been quite a few works that focus on sports events. Thomas et al. \cite{thomas2017} and Wang et al. \cite{Wang:2004:survey} provide detailed survey of some of the current and past works, respectively. Though, Sports-1M, UCF Sports\cite{Soomro2014}, are some of the largest available sports dataset, but they cannot be used to learn models on any one sport in particular, as there is not enough data to recognize all types of events in a single sport. 

We take a large-scale dataset of untrimmed cricket videos, and try to localize the individual cricket strokes being played by the batsman, which is our event of interest, hereafter, referred to as \textit{stroke}. Usually, annotating a dataset of such scale, is done using a crowd sourcing platform like Amazon Mechanical Turk, similar to the works in \cite{Kuehne2011,Heilbron2015,Russakovsky2015}. In this work, we hand annotate only a small subset of validation and test set videos for evaluation purpose and bootstrap the model to make predictions on the entire dataset.

Our motivation for this work is to come up with a generalized solution of developing a large-scale dataset for domain-specific telecast videos. We do not use any form of data(audio, text, etc.) other than the RGB frames of the untrimmed videos, and make minimum assumptions regarding our domain of interest i.e., cricket. Though, there are highly accurate tracking systems, like Hawk-Eye \cite{HawkEye}, which are already being used as Decision Review System (DRS), but their data is private and they have their own set of stationary cameras and sensors installed in the sporting ground. A dataset of such scale can be used to train, for example C3D\cite{Ji2013} type of deep neural networks, and later solve the problem of automatic content-based recognition of types cricket strokes. A direct use of this model would be in automatic commentary generation, apart from content-based browsing and indexing of cricket videos. 

We collected a large set of untrimmed cricket telecast videos, performed a series of pre-processing steps to make them have a uniform frame rate and resolution, and then applied our model for the prediction of temporal localized \textit{stroke} segments. The applied model involves simple shot boundary detection using summed up grayscale histogram difference feature, a couple of camera models that recognize the starting frames of specific camera shots that are part of the \textit{stroke} and finally make \textit{stroke} segment predictions by finding \textit{stroke} shots. 

%

Section~\ref{sec:relatedWork} provides a brief review of the related works. Our methodology is explained in detail in Section~\ref{sec:cricket}, which includes the details about the sub-problems of boundary detection, training of camera models and prediction of \textit{stroke} segments. Section~\ref{sec:experiment} describes our experimental setup and the evaluation metrics used for boundary detection, camera models, and action localization. Finally, we give the results in Section~\ref{sec:results}, followed by the conclusion in Section~\ref{sec:conclusion}.

\section{Related Work}
\label{sec:relatedWork}
The problem of action recognition in videos has picked up pace with the onset of deep neural networks\cite{Karpathy2014,Tran2015,Simonyan2014,Donahue2017,Wang:2014:HAR,Ng2015}. These works modify the architecture of the Convolutional Neural Networks (CNNs) and Recurrent Neural Networks (RNNs) and train them on the video data. As a result, they produce state-of-the-art results at the cost of increasing the number of parameters, and training time. 

The tasks of classification, tracking, segmentation and temporal localization are quite inter-dependent and may involve similar approaches and features. In some works, the problem of temporal action localization has also been tackled using an end-to-end learning network \cite{Wang2016_TSNs,Yeung2015}. Segmenting the object of interest and tracking it in the sequence of frames has been done in a few works like  \cite{Weinzaepfel2016,Jain2014,Jain2017}.

Some of the above approaches need pre-trained models that can be fine-tuned on their own problem specific datasets, while some other use large benchmark datasets for training purpose. Applying such techniques for a sporting event requires a lot of hand-annotated training data, which is hard to get or may include a lot of noise. Automated ways of creating datasets may involve using a third party API, like YouTube Data API (as done in \cite{Karpathy2014}), or extraction using text meta-data of videos, which may not, at all, be accurate. Nga et al.\cite{Nga2014} proposed a method to extract action videos based on the tags. Deciding the relevancy of a tag, in itself, is a research problem. Due to these reasons, content-based action extraction from videos is a good choice for automatic construction of large-scale action dataset. Hu et al.\cite{Hu2011} provide a survey of the content-based methods for video extraction. 

Content-based retrieval of actions from untrimmed videos is tough when the number of actions increase or when a more generalized set of actions is considered. The second problem is prevented in case of sports videos, as the videos of a particular sport have same type of actions performed at intervals. We take cricket as our test case and try to come up with a framework for cricket \textit{stroke} extraction. A cricket \textit{stroke} is a cricketing shot played by a batsman when a bowler bowls a ball. In this paper, we refer to the cricket shot as \textit{stroke}, so that it may not be confused with a video shot, which can be defined as a sequence of frames  captured by a single camera over a time interval during which period it does not switch to any other camera. 

Coming up with a purely content-based domain specific event extraction for untrimmed cricket telecast videos has not been attempted by many. Sharma et al. \cite{Sharma2015} tried to annotate the videos based by mapping the segments to the text-commentary using dynamic programming alignment, which may not always be available or may be noisy. Moreover, their dataset is quite small, as compared to what we are trying to achieve. Some other works have also looked at extraction of cricketing events but they have not considered it at such a scale as ours, or have not tried to generalize their solutions. Kumar et al.\cite{Kumar2014} and some similar works have tried to classify frames based on ground, pitch, or players present in them, or came up with a rule-based solution to classify motion features. None of them have analyzed their results on an entirely new set of cricket videos.   

In our work, we collect a large set of raw untrimmed cricket videos and report our results on a subset of these videos, that have been hand-annotated for the temporal localization task. The architecture (Figure~\ref{fig:architecture}) can be generalized to other sports, since there are no cricket specific assumptions involved. 

Temporal localization using deep neural networks has been quite successful recently, \cite{Zheng2016,Gao_2017_ICCV}. Though, there are other works that do not use deep neural nets, such as Soomro et al. \cite{Soomro_2017_ICCV}.  They use an unsupervised spectral clustering approach to find similar action types and localizing them. Our approach also does not use deep neural networks as our object is to come up with a dataset that is large enough to train CNNs with millions of parameters. Even the pre-trained networks need sufficiently large amount of labeled data for fine-tuning the network. We have done the labeling for only a small set of highlight videos (1GB of World Cup T20 2016) and bootstrapped simple machine learning models trained on only grayscale histogram differences and HOG \cite{N.Dalal2005} features.

\section{Cricket shot extraction}
\label{sec:cricket}
Extracting cricket shots in untrimmed videos is analogous to the action detection (localization) task where the action of interest is a cricket \textit{stroke} being played by a batsman.

\begin{figure}[t]
\centering
\includegraphics[clip, trim=0.5cm 9.5cm 0.5cm 1.5cm, width=1\textwidth]{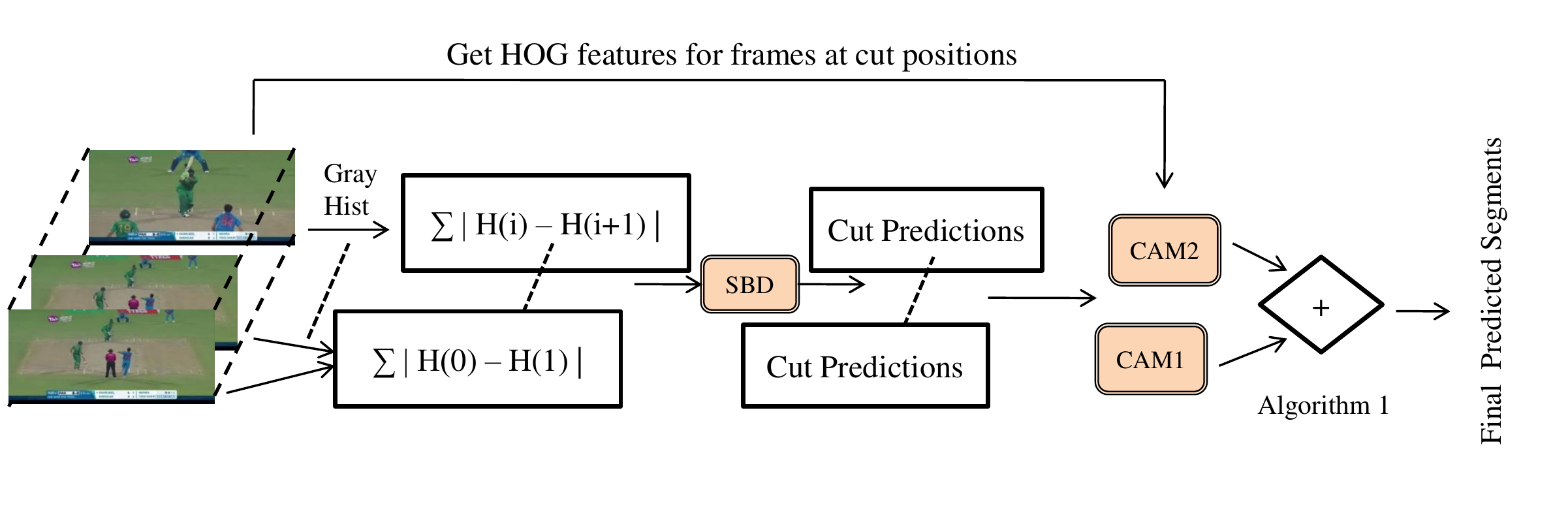}
\caption{Our framework for prediction of cricket \textit{strokes} based on the learned models for shot boundary detection (SBD), camera models (CAM1 and CAM2) and combining the predictions using Algorithm~\ref{algo:naive}}
\label{fig:architecture}
\end{figure}

The live telecast videos of a sports event are created by a set of cameras that have the task of covering most of the sporting arena. There are two types of cameras, fixed and moving. The fixed cameras (assigned to camera-men) have a defined objective of capturing a specific sporting activity, while the moving cameras, for e.g., spider-cams in cricket fields, may be controlled remotely, and may not, necessarily, cover the relevant sporting activity at all times. 

The main sporting activity in a cricket match is a bowler bowling a ball, followed by a batsman playing a \textit{stroke} and then scoring runs. An over consists of 6 such deliveries, each of which is, potentially, an event of interest. Automatically recognizing the outcome of each delivery is a tough problem, which may require a complex system with domain knowledge and learned models that interact in a rule-based manner, as done in \cite{Kumar2014}. In our work, we consider a basic problem of localization of a cricketing event, which is the direction of \textit{stroke} play. Here, the starting point of our event of interest is the camera shot where the bowler is about to deliver the ball, and ending at the camera shot that captures the direction of the cricket stroke played. Generally, our event of interest is captured by one camera shot or at most two subsequent camera shots. An illustration is provided in Figure~\ref{fig:camera_models}, where frame $99$ ($F99$) is the starting point and $F153$ is the ending point of the event of interest. The major portion of the event is captured using two subsequent camera shots, $CAM1$ and $CAM2$\footnote{Please note that we use $CAM1$ and $CAM2$, for referring to the camera shots as well as the models trained on the first frames of these shots}, where $CAM2$ captures a wider area to locate the movement of the ball and then, gradually, focuses on it. There may also be a case where the batsman just taps the ball and it doesn't travel beyond the $CAM1's$ field-of-view. Therefore, we need to segment out either $CAM1$ shots or $CAM1+CAM2$ shots. 

The above type of modeling can be applied (with minor changes) to a number of other sports, like tennis, badminton, or baseball. An illustration of our temporal cricket \textit{stroke} localization model, is given in Figure~\ref{fig:architecture}. The pipeline of simple models, trained on only a small set of sample videos, is used to predict temporal \textit{stroke} segments in a large set of raw untrimmed cricket telecast videos. The evaluation for the bootstrapped predictions is done on a subset of the main dataset, called test set sample. This subset, along with a validation set sample, has been hand-annotated with ground-truth cricket \textit{stroke} segments. An evaluation over these subsets would give an estimate of what we can expect as an overall accuracy. 

A learning-based approach for detection and localization, in order to get generalized results, would require a large amount of labeled data, where labels should contain minimum amount of noise. Such a dataset for cricket telecast videos may help the research community to come up with better learned models for this particular sporting domain. Choosing a purely content-based modeling approach, we need to minimize the manual annotation effort, which leaves us with the idea of bootstrapping smaller models' predictions to the large raw video datasets, where a ``smaller'' model is one having only a few parameters.     

Our method is purely content-based, since, we do not use any extra information, other than the RGB video frames and features extracted from them. The training of simple machine learning models on shallow and high dimensional feature descriptor, for localization, has been performed using a small highlights video dataset. Here, the shallow feature is the summed-up absolute values of histogram differences of consecutive grayscale frames, that are used for shot boundary detection (CUTs), and HOG is the high dimensional feature descriptor of the frame, used to recognize starting frames of CAM1 and CAM2 video shots. The shallow descriptor is, computationally, less expensive as compared to the high dimensional feature descriptor. 
 
The steps involved in the cricket \textit{stroke} extraction are as follows:
\begin{enumerate}
\item Dividing the telecast videos into individual camera shots using shot boundary detection. Here only the CUTs are considered, since the detection of gradual transitions involved an additional overhead. The CUT predictions are done using a random forest\cite{Breiman2001,scikit-learn} model, trained on summed-up absolute values of histogram differences of consecutive grayscale video frames taken from our sample dataset (refer Table~\ref{table:dataset}).
\item A small dataset of first frames was created using only the sample dataset videos for training $CAM1$ and $CAM2$ camera models. These sets had $367$ and $336$ frames each, where nearly half are positive samples and half are negative samples. We trained two linear SVMs\cite{Cortes:1995,scikit-learn} on the HOG \cite{N.Dalal2005} features of the training subset of these first frames. 
\item Having the boundary predictions and the first frame recognition models, a simple approach is to extract only those video shots that give positive results for their first frame HOG features. These will be cricket \textit{strokes}. Algorithm~\ref{algo:naive} describes a simple approach for extracting these events.
\item The evaluation of the extracted cricket \textit{strokes} can be done by defining an evaluation metric and having a test set of human annotated cricket stroke localized segments. We choose the 1D version of IoU metric (Intersection over Union), which is called the temporal IoU (TIoU), and see how much overlap is there between our predictions and the ground-truth annotations.
\end{enumerate}

Given below are the steps in greater detail.

\subsection{Preprocessing}
\label{subsec:preprocess}
The raw cricket telecast videos, collected from different sources, like YouTube and Hotstar, had different frame rates and resolutions. All the videos were resized to $360 \times 640$ with a constant frame rate of $25$FPS. This step was carried out using FFmpeg. Having a constant frame rate and frame dimensions ensures the uniformity of the motion features that may be extracted from the videos.

\subsection{Shot boundary detection}
\label{subsec:sbd}
Shot boundary detection has been studied for decades and is, often, the first step of any content-based video processing system. There may be two types of boundaries, such as a hard CUT transitions (occurs where one camera shot ends abruptly and the next begins) or gradual transition (like fade, wipe, dissolve, etc.). Our focus is only on the detection of CUT boundaries, since, CUTs are the most common in sports telecast videos. As a future work, one might focus on detection of gradual transitions as well, but that would tend to increase the overall processing time, which needs to be minimized while dealing with any kind of large-scale data processing. The use of CUT predictions, is when we need to jump directly from one camera shot to the next, in an untrimmed video. Iterating over the CUT predictions, and extracting the HOG features from only the first frames, speeds up our processing. 

We tested for histogram differences (grayscale and RGB) and weighted-$\chi^2$ differences of consecutive frames. The equation~\ref{eqn:histdiffs} gives the value of summed-up absolute value of histogram differences for consecutive $i^{th}$ and $(i+1)^{th}$ grayscale frames, where $N$ is the number of bins representing the different gray-levels.  

\begin{align}
\label{eqn:histdiffs}
	\begin{split}
	 D(i, i+1) = \sum\limits_{n=1}^{N} \mid ( H_{i}(n)-H_{i+1}(n) ) \mid    	  
	\end{split}
\end{align}



\subsection{Camera Models}
\label{subsec:cam_models}
\begin{figure}
\centering
\includegraphics[clip, trim=0.5cm 7.5cm 0.5cm 1.5cm, width=0.80\textwidth]{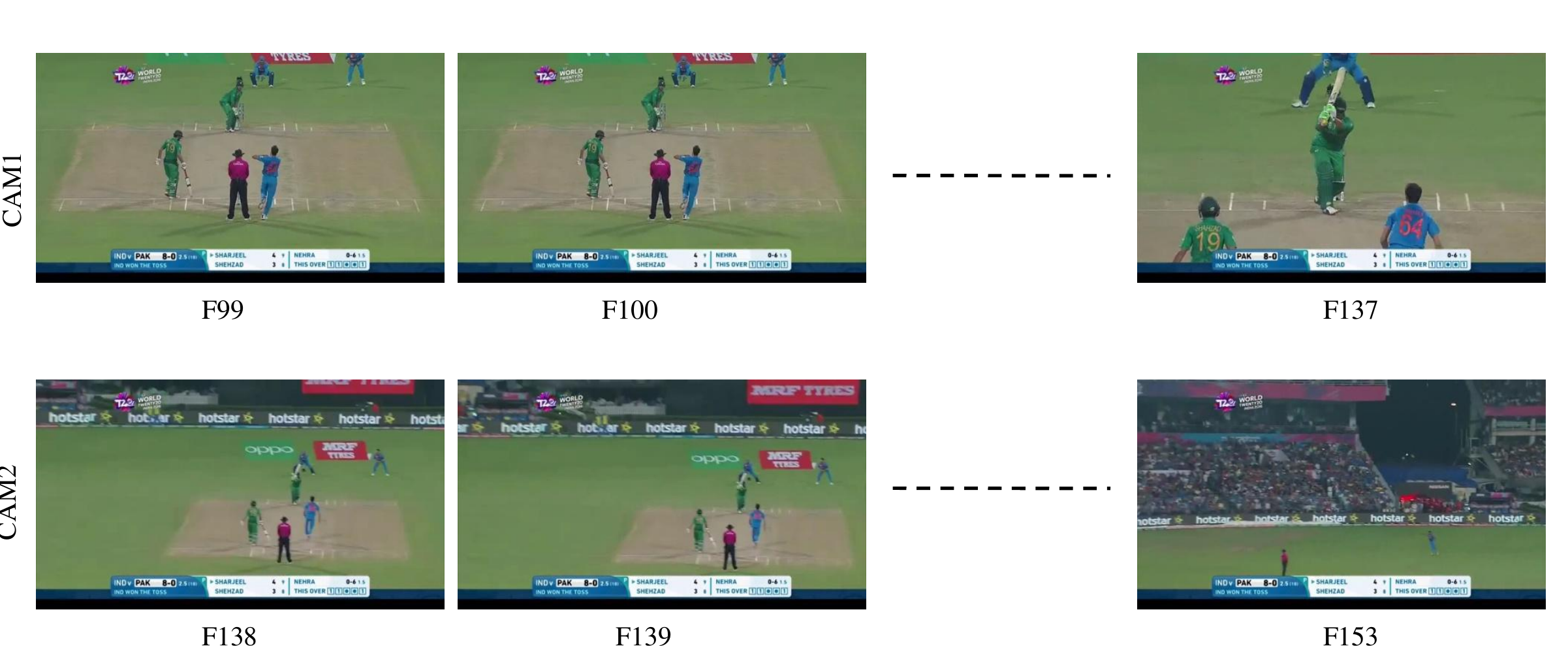}
\caption{The sequence of frames in a cricket \textit{stroke}. As the ball goes outside the field-of-view of $CAM1$, the camera is switched to $CAM2$. Note there is a CUT between frames $F137$ and $F138$}
\label{fig:camera_models}
\end{figure}

We may assume that each fixed camera has a defined task, i.e., it will, regularly, cover similar actions being performed. The starting scene for cricket will have a bowler taking a run-up to the bowling crease and the batsman standing at the other end of the pitch ready to face the delivery. This \textit{stroke} can be detected by identifying the first frame of this video shot (using predicted CUT positions) and segmenting out the shot till the next CUT position. The accuracy of this method relies on how accurate the CUT predictions are and how accurately we detect the first frames of the \textit{stroke}. Figure~\ref{fig:camera_models} shows a sample of cricket \textit{stroke} containing $CAM1$ and $CAM2$ video shots.

\subsection{Extraction Algorithm}
\label{subsec:naive}
Algorithm~\ref{algo:naive} is a simple approach to localize our activities of interest, given the individual pretrained camera models and a set of cut predictions for an untrimmed video. The final predictions for the \textit{stroke} segments may be of any length, which includes segments obtained from a false positive CUT followed by a false positive $CAM1$/$CAM2$ prediction. This occurs, mostly, at places where there are gradual transitions. These false positive segments, are of small duration, and can, simply, be neglected by a filtering step, which is explained in Section~\ref{subsec:filtering}. 
\begin{algorithm}
\caption{Extract cricket strokes}\label{algo:naive}
\begin{algorithmic}[1]
\Procedure{localizeActions}{$srcVideo, videoCuts, cam1, cam2$}
\Comment{$srcVideo :$ the input video}
\Comment{$videoCuts :$ list of predicted CUTs for $srcVideo$}
\Comment{$cam1, cam2 :$ trained camera models}
\State $\textit{vid\_segments, start\_frame, end\_frame} \gets emptyList(\left[ \; \right]), -1, -1 $ 
\For{$i,cut \gets enumerate(videoCuts)$}	\Comment{Iterate over the CUT positions}
	\State $ \text{set} \; \textit{cut} \; \text{position in} \; \textit{srcVideo} $
	\State $frame \gets readFrameFromVideo()$	
	\State $hog \gets computeHOG(frame)$
	\State $cam1\_pred \gets cam1.predict(hog)$
	\State $cam2\_pred \gets cam2.predict(hog)$	
	\If {$start\_frame = = -1$}
		\If {$cam1\_pred = = 1$}		\Comment{positive prediction for $cam1$}
		\State $start\_frame \gets cut$
		\EndIf
	\ElsIf {$start\_frame \ge 0$}
		\If {$cam2\_pred = = 0$}
			\State $end\_frame \gets cut - 1 $
			\State $vid\_segments.append(\; \left[ start\_frame, \; end\_frame \right] \;)$ \Comment {save a predicted segment to list}
			\State $start\_frame \gets end\_frame \gets -1$
			\If {$cam1\_pred = = 1$}
				\State $start\_frame \gets cut$ 
			\EndIf
		\Else
			\If {$(i+1) < \text{len}(videoCuts)$}
				\State $end\_frame \gets videoCuts[i+1] - 1$
			\Else
				\State $end\_frame \gets \text{nFrames}(srcVideo) - 1$ \Comment {get number of frames in video}
			\EndIf
			\State $vid\_segments.append(\; \left[ start\_frame, \; end\_frame \right] \;)$			
			\State $start\_frame \gets end\_frame \gets -1$			
		\EndIf
	\EndIf
\EndFor
\State $\text{return} \; vid\_segments $
\EndProcedure
\end{algorithmic}
\end{algorithm}

\subsection{Bootstrapping the predictions}
\label{subsec:bootstrapping}
We refer to the final predictions made on the large dataset, as the bootstrapped predictions, since, the trained models used an entirely different dataset (Highlight videos) for training. The accuracy of these predictions depends, largely, on the Highlight videos.

\section{Experimentation Details}
\label{sec:experiment}

This section describes our experimental setup in detail.

\subsection{Data Description}
\label{subsec:data}
The raw cricket telecast videos were first transformed to have a constant frame rate (25 FPS) and resolution ($360 \times 640$) by using FFmpeg. We had a sample dataset that had only 26 highlight videos, each of around 2-5 minute duration. This dataset was set aside for experimentation and training of intermediate CUTs and CAM models. The main dataset of untrimmed videos (also referred to as \textit{full} dataset) had 273 GB ($>73$ million frames) of untrimmed videos from various sources, containing Test Match videos, ODI videos and T20 videos. We neglected any local cricket match videos, where the telecast camera was positioned at a non-standard position. A brief summary of both the datasets is given in Table~\ref{table:dataset}. Further, we partitioned the datasets into training, validation and testing sets in the given ratios. The highlight videos are annotated with CUT positions and \textit{stroke} segments. A subset of validation set videos and a  subset of test set videos of the \textit{full} dataset are manually annotated with the \textit{stroke} segments. They are used for the final evaluation.

\setlength{\tabcolsep}{5pt}
\begin{table}
\begin{center}
\caption{Details of our datasets}
\label{table:dataset}
\begin{tabular}{lll}
\hline\noalign{\smallskip}
Property & Highlights dataset & Full dataset\\
\noalign{\smallskip}
\hline
\noalign{\smallskip}
No. of Videos  		& $26$ 			& $1110$ \\
Total Size 			& $1GB$ (approx.) & $273GB$ (approx.) \\
Dimensions (H,W) 	& $360 \times 640$ & $360 \times 640$ \\
Frame Rate (FPS) 	& $25.0$ 		& $25.0$ \\
Event Description 	& ICC World Cup T20 2016 & Varied sources \\
Training set 		& 16 videos ($\approx60\%$)	& $\approx50\%$ \\
Validation set 		& 5 videos ($\approx20\%$)	& $\approx25\%$ \\
Test set 			& 5 videos ($\approx20\%$)	& $\approx25\%$ \\
Annotations			& CUTs, \textit{strokes} 	& \textit{strokes} on a subset  \\
\hline
\end{tabular}
\end{center}
\end{table}

\subsection{Shot Boundary Detection}
\label{subsec:sbd:exp}
We tested a number of approaches for detecting the shot boundaries (CUTs) based on only the sum of the absolute values of histogram differences of consecutive frames; like global thresholding on grayscale/ RGB differences \cite{Smeaton2010}, weighted-$\chi^2$ differences \cite{Ko2006} and applying simple classifiers on these features. The best performance on the sample datasets' test set was given by applying a random forest model on grayscale histogram differences of consecutive frames \ref{sec:results}. 

\textbf{Evaluation Criteria:} We follow the evaluation process of TRECVid for detection of only CUT transitions. Details are specified in \cite{Ruiloba1999}. A false positive is referred to as an \textit{insertion}, while a false negative to a \textit{deletion}. The presence of gradual transition or setting a low threshold (in case of global thresholding), creates \textit{insertions}, and, as a result, tends to reduce the overall precision. While setting a global threshold to a high value misses out on actual CUT boundaries and reduces the overall recall. Therefore, F-score is a suitable evaluation criteria for CUT boundary detection.

\subsection{Camera Models}
\label{subsec:cam_models:exp}
\begin{figure}
\centering
\includegraphics[clip, trim=0.5cm 1.5cm 1.5cm 1.5cm, width=0.90\textwidth]{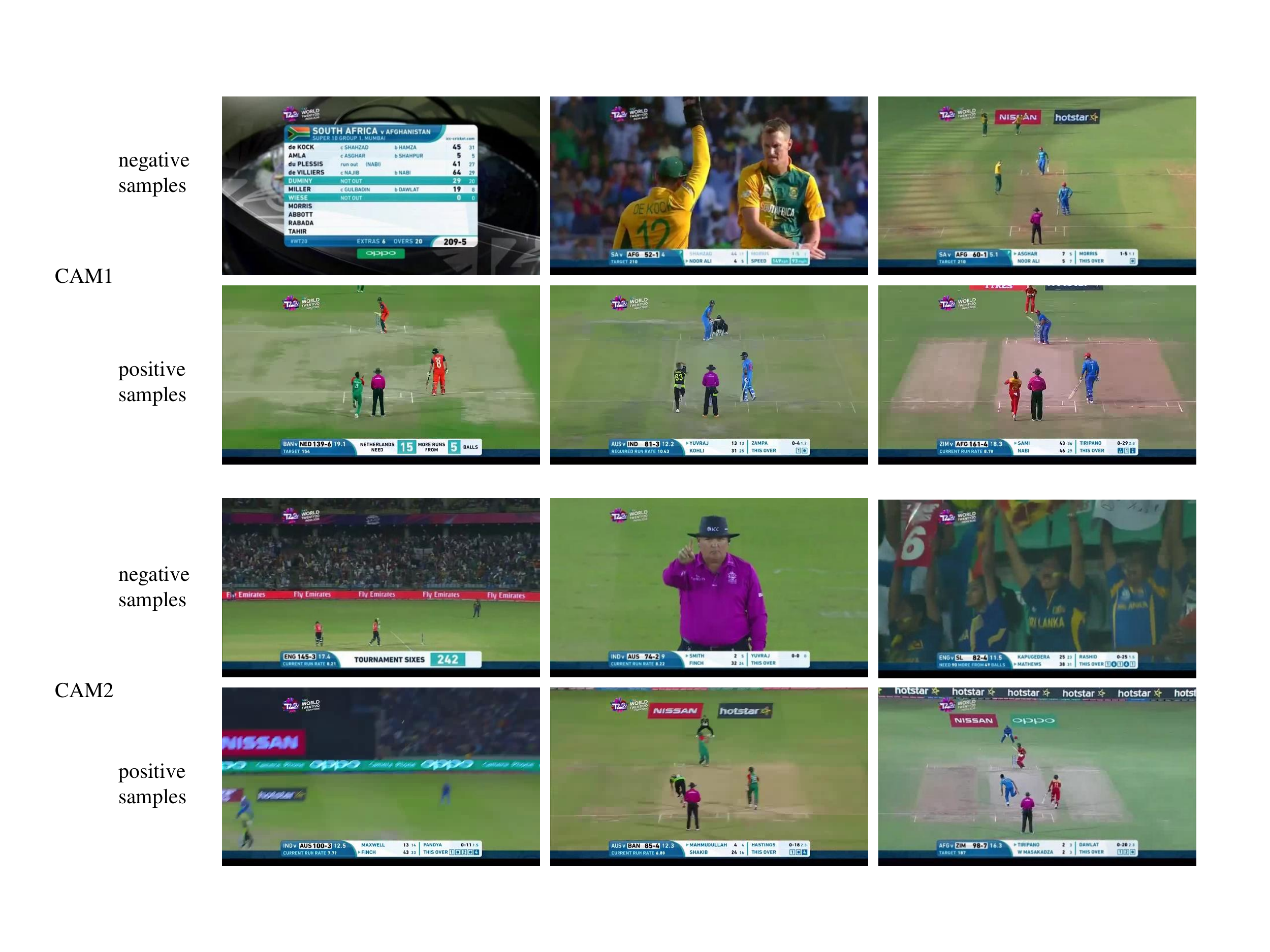}
\caption{A few training examples from our $CAM1$ and $CAM2$ samples datasets}
\label{fig:camsamples}
\end{figure}

The only assumption in our work that is specific to cricket, as illustrated in figure~\ref{fig:camera_models}, is that $CAM1$ and $CAM1$+$CAM2$ video shots comprise of our events of interest, therefore, need to be localized. These video shots can be recognized by extracting a number of features, such as shape features, tracking cricketing objects, textures, etc, but all of these may not be generally applicable to other sporting domains. We notice that the first frames of $CAM1$ shots are quite ``similar'' and same is the case with first frames of $CAM2$. We chose to extract a fine grained HOG feature vector for the grayscale first frames of $CAM1$ and $CAM2$ and train simple machine learning models on them. The sample datasets for these two models had $367$ and $336$ frames, respectively, that include about half positive samples and half negative samples. These samples were extracted manually from a subset of highlight sample videos dataset and the negative frames are the first frames of some other random camera shots that are not cricket \textit{strokes}. Figure~\ref{fig:camsamples} shows a few training frames used for training the CAM models. 

\subsection{Filtering final predictions}
\label{subsec:filtering}
The final predictions are of the form $[s_i, e_i]$ where $s_i$ and $e_i$ are the starting and ending frame positions, respectively, for the $i^{th}$ predicted segment. The value of $d_i=e_i-s_i$ for a cricket stroke should be sufficiently large, considering the fact that the actions take time and the frame rate is constant at $25$FPS. A low value of $d_i$, generally, occurs, mainly, due to the gradual transitions, fast camera motion, or some advertisements occurring in between the telecast video. We apply a filtering step to remove any segments that have $d_i < T$. The results for $T=\lbrace0, 10, 20, ..., 100\rbrace$ on the validation set samples, are given in Figure~\ref{fig:tiouFilter}. As the best result occurs for $T=60$, therefore, we set this value for our final accuracy, on the test set samples.

\subsection{Temporal Intersection over Union (IoU)}
\label{subsec:tiou}
The metric used for evaluation of the localization task is the Weighted Mean Temporal IoU, motivated from \cite{Baraldi:2015}. If for a specific untrimmed video, the set of predicted segments are $S=\left\lbrace s_1, s_2, ..., s_m \right\rbrace$ and the set of ground truth segments are $\hat{S}=\left\lbrace \hat{s}_1, \hat{s}_2, ..., \hat{s}_n \right\rbrace$, then the Weighted Mean Temporal IoU is given by equation~\ref{eqn:meantiou}, where mean is taken over the weighted sum of $TIoU_i$ values of $i^{th}$ untrimmed video, each weighted by the number of ground truth segments $n_i$ in that video and $V$ being the total number of untrimmed videos in the test dataset. Each segment $s \in S$ or $\hat{s} \in \hat{S}$ is of the form \textit{[start segment position, end segment position]}. In equation~\ref{eqn:vidtiou}, $TIoU_{i}$ is calculated by taking the maximum overlaps of each ground truth segment with all the predicted segments and vice versa. 

\begin{align}
\label{eqn:meantiou}
\begin{split}
	 TIoU_{mean} = \frac{ \sum_{i=1}^{V} n_i  \; TIoU_i}{\sum_{i=1}^{V} n_i}
\end{split}
\end{align}

\begin{align}
\label{eqn:vidtiou}
\begin{split}
	 TIoU_{i} = \frac{1}{2} \left[ \frac{1}{n} \sum_{i=1}^{n} \max_{j \in N_m}  \; \frac{s_j \cap \hat{s_i}}{s_j \cup \hat{s_i}} + \frac{1}{m} \sum_{i=1}^{m} \max_{j \in N_n}  \; \frac{s_i \cap \hat{s_j}}{s_i \cup \hat{s_j}} \right]
\end{split}
\end{align}


\subsection{A Note on Data Parallelism}
\label{subsec:parallelism}
When dealing with any large-scale data processing, parallelism is essential. The extraction of a single feature from the entire dataset may take weeks, if performed serially. For our $full$ dataset of $\approx 273$ GB, the extraction of grayscale histogram differences takes more than 10 days. If we consider extraction of any fine grained feature like HOG, which is computationally expensive, that would take much more than time. We followed a data parallelism approach for any kind of feature extraction or prediction. The untrimmed videos were divided into batches and each batch was parallelized over a fixed number of processes running in parallel over different cores. Table~\ref{table:time} shows approximate time for some of the feature extraction operations, with different batch sizes and running on ``\#Jobs'' number of cores in parallel. 

\setlength{\tabcolsep}{5pt}
\begin{table}
\begin{center}
\caption{Execution times compared with and without data parallelization. The Sorted Videos column refers to whether the videos were sorted based on their sizes. }
\label{table:time}
\begin{tabular}{llllll}
\hline\noalign{\smallskip}
Feature & Sorted Videos? & Batch Size  & \#Jobs & Time(approx.) \\
\noalign{\smallskip}
\hline
\noalign{\smallskip}
HDiffs Gray & Unsorted & $1$ & $1$  &  $>240$ hours \\
HDiffs BGR  & Unsorted & $50$  & $10$  & $>40$ hours \\
HDiffs BGR & Sorted & $50$ & $10$  &  $6.62$ hours \\
Wt-$\chi^2$Diffs & Sorted & $50$ & $10$  & $6.66$ hours \\
\hline
\end{tabular}
\end{center}
\end{table}

\section{Results and Discussion}
\label{sec:results}
In this section, we describe the results of our individual models and our final predicted segments. 

\textbf{Shot Boundary Detection:} CUTs detection worked best with the grayscale histogram difference feature. The number of histogram bins was set to maximum (256), since changing it to any lower value decreased the accuracy. In case of global thresholding, weighted-$\chi^2$ difference values, performs better than the grayscale histogram differences and RGB histogram differences. The accuracy for a trained random forest model exceeds the global thresholding approach by a big margin, and a random forest trained on summed-up absolute grayscale histogram difference features, is works better than the one trained on weighted-$\chi^2$ differences. Refer Table~\ref{table:evalsbd} for the accuracy of the CUTs model on the test set of our Highlight videos dataset.  
\setlength{\tabcolsep}{5pt}
\begin{table}
\begin{center}
\caption{Evaluation results for shot boundary detection model}
\label{table:evalsbd}
\begin{tabular}{llllll}
\hline\noalign{\smallskip}
Model & Feature & Dataset & Precision & Recall & F-score\\
\noalign{\smallskip}
\hline
\noalign{\smallskip}
\textbf{SBD RF(Bins:256)} & HDiffs(Gray) & Highlights  & $0.9796$  &  $0.9711$ & $0.9753$ \\
SBD RF(Bins:128) 		  & HDiffs(Gray) & Highlights  	& $0.9380$  &  $0.8510$ & $0.8924$ \\
\hline
\end{tabular}
\end{center}
\end{table}

\textbf{Camera Models:} We train linear SVMs on the HOG features of the first frames. The accuracy values for the two trained models is given in table~\ref{table:evalcam}. Out of 111 test samples, there was 1 false negative and 1 false positive for $CAM1$, while for $CAM2$, there were 4 false positives and 1 false negative. For final predictions, we used these models, on extracted HOG features of the first frames. 
\setlength{\tabcolsep}{5pt}
\begin{table}
\begin{center}
\caption{Evaluation results for CAM models.}
\label{table:evalcam}
\begin{tabular}{llll}
\hline\noalign{\smallskip}
Model & Feature  & \#Test samples & Error \\
\noalign{\smallskip}
\hline
\noalign{\smallskip}
\textbf{CAM1 LinearSVM} & HOG & $111$  &  $1.80 \%$ \\
\textbf{CAM2 LinearSVM} & HOG & $101$  & $4.95 \%$ \\
\hline
\end{tabular}
\end{center}
\end{table}

\begin{figure}
\centering
\includegraphics[clip, width=0.60\textwidth]{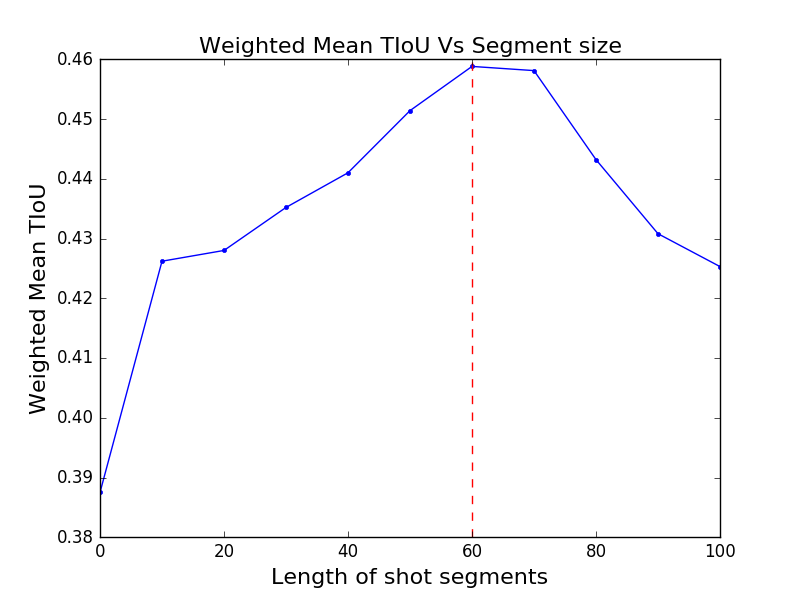}
\caption{Evaluation on validation set samples after filtering out segments less than given size }
\label{fig:tiouFilter}
\end{figure}

\textbf{TIoU:} The weighted mean TIoU metric can be applied to any temporal localization task for untrimmed videos. A minimum value of $0$ for this metric denotes that our predictions are completely off, while a the maximum value of $1$ denotes that we predict perfectly. Our predictions of the \textit{stroke} segments were filtered, as explained in section~\ref{subsec:filtering}, and then evaluated on the validation set sample videos.  Refer to figure~\ref{fig:tiouFilter} for the results on the validation set samples. The value of the filtering parameter was set to $60$, i.e., any predicted shot segment that is of size less than $60$ will be neglected. The final evaluation result is calculated against the $30$ untrimmed videos taken from the test set of our \textit{full} dataset. The weighted mean TIoU was \textbf{0.5097}.

We tried to modify the algorithm~\ref{algo:naive} to make predictions on the first 5 frames and take only those CAM shots for which the $p$ out of 5 are positive predictions. But, in each case, the accuracy was lower than what has been reported.

\section{Conclusion and Future work}
\label{sec:conclusion}

In this work, we demonstrated a simple approach for extraction of similar sporting actions from telecast videos by performing experiments over a large-scale dataset of cricket videos. Here, our action of interest was a cricket \textit{stroke} played by a batsman. The sequence of extraction steps, that we followed, may be generalized for other sporting events as well. Extracting cricket strokes is a temporal localization problem, for which the final accuracy (weighted mean TIoU) was \textbf{$0.5097$} which is quite accurate, considering the fact that we did not use any complicated approach, such as training CNNs or extracting any motion information. 

Our objective is to come up with a large-scale cricket actions dataset, which can be used to train deep neural networks for cricket video understanding. There is a lot of scope for improvement of our results, which is our future work. In addition to that, if we are given the individual cricket \textit{stroke} segments, how we can cluster them into different types, by looking into the motion features. These clusters should represent the different types of cricketing \textit{strokes}, such as, stroke towards mid-wicket, towards long-off etc.  

\bibliographystyle{splncs04}
\bibliography{mybibfile}
\end{document}